\documentclass[10pt, a4paper]{article}

\usepackage[final]{lrec2026} 

\usepackage[utf8]{inputenc}
\usepackage[T5]{fontenc}
\usepackage[vietnamese,english]{babel}
\usepackage{tabularx}
\usepackage{ragged2e}
\usepackage{amsmath,amssymb}
\usepackage{graphicx}
\usepackage{caption}
\usepackage{subcaption}
\usepackage{textcomp}
\usepackage{hyperref}
\usepackage{booktabs}
\usepackage{multirow}
\usepackage{enumitem}
\usepackage{xcolor}
\usepackage{fvextra} 
\usepackage[T1,T5]{fontenc} 
\usepackage[utf8]{inputenc}

\newcommand{\aff}[1]{\textsuperscript{#1}}
\newcommand{\aname}[2]{#1\aff{#2}}

\title{ViMedCSS: A Vietnamese Medical Code-Switching Speech Dataset \& Benchmark}

\name{
\parbox{\textwidth}{\centering
{\bfseries\large
\aname{Tung~X.~Nguyen}{1,2},
\aname{Nhu~Vo}{1,4},
\aname{Giang-Son~Nguyen}{1,2},
\aname{Duy~Mai~Hoang}{3},\\
\aname{Chien~Dinh~Huynh}{3},
\aname{I\~nigo~Jauregi~Unanue}{4},
\aname{Massimo~Piccardi}{4},
\aname{Wray~Buntine}{1},
\aname{Dung~D.~Le}{1,2}
}}
}

\address{
\aff{1} College of Engineering and Computer Science, VinUniversity, Vietnam\\
\aff{2} Center for AI Research, VinUniversity, Vietnam\\
\aff{3} College of Health Sciences, VinUniversity, Vietnam\\
\aff{4} University of Technology Sydney, Australia\\
\{tung.nx, nhu.vd, son.ng, duy.hm, chien.hd, wray.b, dung.ld\}@vinuni.edu.vn\\
\{DiepNhu.Vo, Inigo.JauregiUnanue, Massimo.Piccardi\}@uts.edu.au
}

\abstract{
Code-switching (CS), which is when Vietnamese speech uses English words like drug names or procedures, is a common phenomenon in Vietnamese medical communication. This creates challenges for Automatic Speech Recognition (ASR) systems, especially in low-resource languages like Vietnamese. Current most ASR systems struggle to recognize correctly English medical terms within Vietnamese sentences, and no benchmark addresses this challenge. In this paper, we construct a 34-hour \textbf{Vi}etnamese \textbf{Med}ical \textbf{C}ode-\textbf{S}witching \textbf{S}peech dataset (ViMedCSS) containing 16,576 utterances. Each utterance includes at least one English medical term drawn from a curated bilingual lexicon covering five medical topics. Using this dataset, we evaluate several state-of-the-art ASR models and examine different specific fine-tuning strategies for improving medical term recognition to investigate the best approach to solve in the dataset. Experimental results show that Vietnamese-optimized models perform better on general segments, while multilingual pretraining helps capture English insertions. The combination of both approaches yields the best balance between overall and code-switched accuracy. This work provides the first benchmark for Vietnamese medical code-switching and offers insights into effective domain adaptation for low-resource, multilingual ASR systems. 
 \\ \newline  }

\begin{document}

\maketitleabstract
\section{Introduction}

Code-switching (CS) is pervasive in Vietnamese medical communication, where English clinical terms (drug names, procedures, biomarkers) appear within otherwise Vietnamese utterances. Prior work across languages shows that ASR errors peak precisely on the embedded-language portions of an utterance—i.e., at the points where non-matrix terms are inserted—highlighting the need for language-tagged, degree-controlled evaluation to diagnose model behavior on these spans \cite{seame,decm,agro2025code}. Yet there is no open benchmark centered on Vietnamese medical CS that simultaneously supports a systematic study of practical remedies, from injecting domain term lists during decoding to CS-oriented adaptation on modern encoder–decoder backbones. 

Clear and accurate medical communication is a cornerstone of patient safety and clinical effectiveness \cite{sharkiya2023quality}. In Vietnam, where healthcare professionals frequently alternate between Vietnamese and English medical terminology during consultations, lectures, and patient education, this code-switching reflects the globalization of medicine but also introduces a significant risk of misunderstanding \cite{chen2025code}. Misrecognition of key terms—such as drug names, anatomical structures, or diagnostic procedures—by automated transcription systems can lead to errors in clinical documentation, medication administration, and data reporting. In medical education and research, inaccurate recognition of bilingual terminology diminishes the clarity of lectures and assessment materials, undermining both comprehension and patient-care competence among trainees \cite{hamad2025decolonizing}. A reliable system for detecting and transcribing code-switched speech is therefore not merely a technical goal but a public-health necessity. It ensures that digital records accurately capture the clinician’s intent, supports high-quality medical training materials, and facilitates inclusive communication with non-specialist audiences and multilingual patients.

Model capacity and pretraining have rapidly advanced Vietnamese ASR, spanning both multilingual architectures \cite{radford2023robust, pratap2024scaling} and Vietnamese-optimized variants \cite{PhoWhisper,Thai_Binh_Nguyen_wav2vec2_vi_2021,zhuo25}. Across these families, a characteristic trade-off emerges in code-switching: models optimized for Vietnamese tend to reduce sentence-level errors in the matrix language, while broadly trained multilingual models better recognize embedded English segments. A benchmark that explicitly separates overall accuracy from accuracy on code-switched spans is therefore needed.

We introduce ViMedCSS,\footnote{ \texttt{\url{https://huggingface.co/datasets/tensorxt/ViMedCSS}}} a Vietnamese medical code-switching speech dataset in which every utterance contains at least one code-switched medical term drawn from a bilingual lexicon. The corpus comprises 34.57 hours and 16{,}576 utterances across five topics, and includes a held-out hard split of rare/unseen terms to test generalization beyond the training vocabulary. We establish zero-shot baselines with state-of-the-art multilingual and Vietnamese ASR systems, then systematically compare fine-tuning on a Whisper-based Vietnamese backbone across complementary approaches to code switching—most notably contextual biasing during decoding versus language-identity–guided adaptation—together with parameter-efficient adapters, post-decoding normalization, and their hybrids. Our evaluation separates overall accuracy from performance on code-switched spans metrics, yielding practical guidance on which strategies most effectively handle medical code switching in Vietnamese ASR.

\section{Related Work}
\label{sec:relatedwork}

Vietnamese ASR has benefited from both large multilingual pretraining and targeted Vietnamese adaptation. Whisper provides strong multilingual zero-shot performance and serves as a widely used encoder–decoder baseline \cite{radford2023robust}, while PhoWhisper adapts the same architecture to Vietnamese via fine-tuning on an 844\,h corpus covering diverse speakers and styles \cite{PhoWhisper}. MMS scales wav2vec~2.0 \cite{wav2vec2} to over one thousand languages with competitive CTC baselines \cite{pratap2024scaling}. On the monolingual side, wav2vec2-base-vi leverages large-scale unlabeled YouTube audio and is fine-tuned on VLSP labels \cite{Thai_Binh_Nguyen_wav2vec2_vi_2021}, and VietASR employs a Zipformer encoder with ASR-biased self-supervision, pre-trained on roughly 70k\,h and fine-tuned on 50\,h of labeled Vietnamese speech \cite{zhuo25}. Public Vietnamese resources such as VIVOS and multilingual corpora like FLEURS further support training and evaluation \cite{luong_2016_7068130,fleurs}. Together these systems span key design axes—multilingual vs.\ Vietnamese-only training, encoder–decoder vs.\ CTC decoding, and compact vs.\ large capacity—and constitute the primary baselines against which we study Vietnamese medical code switching.

Privacy constraints limit open medical speech corpora. For Vietnamese, VietMed provides a mix of labeled and large unlabeled medical audio with ASR baselines and recipes \cite{vietmed}. For multilingual clinical communication that includes Vietnamese, MultiMed-ST offers a large many-to-many medical speech–translation corpus with analyses that also examine code-switching phenomena \cite{le2025multimed}. On the text side, MedEV introduces a sizeable Vietnamese–English medical parallel corpus and benchmarks multiple Machine Translation (MT) systems, showing clear gains from domain-specific fine-tuning \cite{medev}. Together, these efforts advance Vietnamese medical ASR and MT, but none target systematic evaluation of code-switched medical terminology within ASR or the role of contextual biasing, which motivates our benchmark.

A recent review synthesizes datasets, metrics, and modeling patterns for end-to-end CS ASR, emphasizing language-split reporting and CS-degree slicing \cite{agro2025code}. Canonical testbeds include SEAME, with time-aligned language boundary tags \cite{seame}, and the ASRU 2019 Mandarin–English challenge \cite{asru2019}, while DECM contributes a German–English evaluation set with word-level tags and explicit low/mid/high CS bins \cite{decm}. CS-FLEURS broadens with a benchmark spanning 52 languages and over one hundred code-switched pairs in general domain \cite{yan25c_interspeech}. On the modeling side, Whisper-based adaptations that leverage language identity (LID) have proven effective: attention-guided, parameter-efficient finetuning selects and steers LID-sensitive heads \cite{aditya2024attention}, and complementary work refines Whisper via encoder improvements and language-aware decoding \cite{zhao2025adapting}.

Beyond architectural adaptation, contextual biasing targets rare, domain terms at inference. Neural–symbolic approaches such as TCPGen integrate a prefix-trie of bias words into end-to-end decoders and reduce errors on long-tail entities \cite{sun2023can} To handle large catalogs, ranking/selection methods forward only the top-$k$ most relevant items to the decoder \cite{hou25_interspeech}. Dynamic vocabulary further injects bias entries as single tokens on the fly, avoiding heavy external Language Models (LMs) or rescoring \cite{sudo2024contextualized, sudo25b_interspeech}. 

\section{Vietnamese Medical Code-Switching Speech dataset - ViMedCSS}
\label{sec:append-how-prod}

\subsection{Construction}
\begin{figure*}[t]
  \centering
  \includegraphics[width=\linewidth]{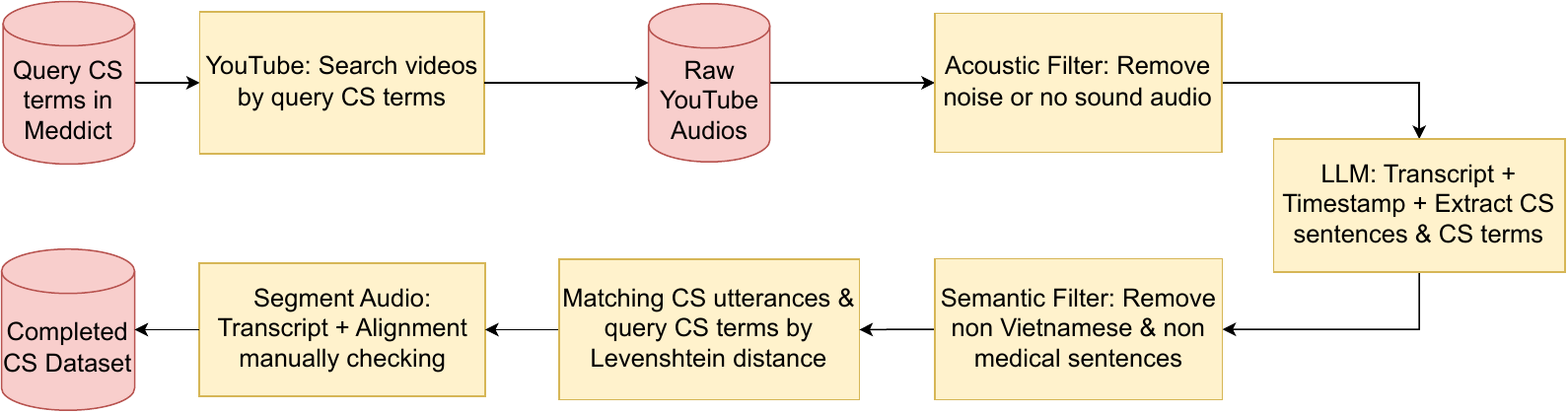}
  \caption{Dataset construction pipeline for Vietnamese medical code-switching.}
  \label{fig:pipeline_cs}
\end{figure*}

 As seen from Figure \ref{fig:pipeline_cs}, we start the dataset construction pipeline from the \textit{Meddict}\footnote{\url{https://meddict-vinuni.com/}} dictionary, an English–Vietnamese medical lexicon created at VinUniversity with 64{,}232 entries. Meddict offers curated translations of specialized clinical terminology and is protected under Intellectual Property Rights Certificate No.~3365/2024/QTG. It is released for academic and healthcare use under the institution’s license. From this bilingual source, we select entries whose Vietnamese usage retains an English or foreign-root surface form; these define our set of code-switched (CS) medical terms. In total, we extract 3{,}203 CS terms from the dictionary.

Using these terms as queries, we retrieve Vietnamese medical videos from public platforms. Candidates must have Vietnamese titles and belong to the medical domain; both conditions are automatically checked with a large language model (LLM). We crawl more than 13{,}000 YouTube videos and discard items with music-only or non-speech audio before transcription. Each remaining audio track is transcribed with Gemini~2.5~Pro to produce time-aligned text and to automatically flag candidate CS sentences and terms. In aggregate, over 700 hours of audio are processed in this step.

We provide the LLM with the following instruction to obtain sentence-level timestamps and CS spans in a machine-readable format:

\begin{quote}
\small
\textbf{Transcription Task:} You are an advanced transcription assistant for Vietnamese medical audio. Do the following:
\begin{enumerate}[nosep,leftmargin=*,label=\arabic*)]
    \item Transcribe the Vietnamese speech.
    \item Segment into sentences of approximately 5--15 seconds.
    \item For each segment, output \texttt{start\_time}, \texttt{end\_time}, and \texttt{text}.
    \item Detect segments that contain any non-Vietnamese terms (e.g., English, Chinese, technical product/brand names).
    \item Return the complete set of segments, and separately the subset of code-switch segments. For each code-switch segment, list the non-Vietnamese terms that appear.
\end{enumerate}

All output must be a single valid JSON object with keys:
\begin{itemize}[nosep,leftmargin=1.2em]
    \item \texttt{"segments"}: an array of all segment objects \{\texttt{start\_time}, \texttt{end\_time}, \texttt{text}\}.
    \item \texttt{"code\_switch\_segments"}: an array of code-switch segment objects \{\texttt{start\_time}, \texttt{end\_time}, \texttt{text}, \texttt{cs\_term}: [...]\}.
\end{itemize}
Return only JSON; do not include any additional prose.
\end{quote}

We then apply LLM-assisted \emph{semantic filtering} to remove utterances that are non-Vietnamese or off-domain, and we \emph{normalize} surface forms to a canonical dictionary (orthography, hyphenation, casing, common variants) to ensure consistent term identity across transcripts. Because the CS spans returned by the LLM may not exactly match queried dictionary entries, we further align terms by computing the Levenshtein distance between each dictionary item and each sentence, assigning the closest canonical entry to the detected span. After this filtering and normalization pass, a little over 34 hours of audio remain as valid Vietnamese medical CS data.

Finally, we segment the raw audio into 3–29\,s utterances and perform manual alignment checks for quality control. The resulting corpus is domain-focused and guarantees at least one CS medical term per utterance, enabling evaluation along both contextual-biasing and code-switching dimensions.

\subsection{Sampling and Quality Verification}
To assess annotation reliability, we sampled 500 utterances (approximately one hour) from the 34.57-hour corpus, stratified across the five topics to preserve domain diversity. Two trained annotators independently reviewed each utterance following the project guidelines, assigning labels for (i) transcription errors on Vietnamese words and (ii) errors on code-switched terms. 

Inter-annotator agreement, measured with Cohen’s kappa \cite{cohen1960}, was $\kappa=0.65$, indicating substantial consistency. This suggests that the guidelines were clear and that the sampled set is representative of the broader corpus.

The main source of discrepancy arose from imperfect segment boundaries: timestamps were occasionally misaligned, making end points ambiguous. We therefore added an automatic boundary-refinement step to the pipeline (timestamp smoothing and alignment correction) before downstream processing, which reduced these mismatches in subsequent audits.

\subsection{Statistics}
\label{stat}
Table~\ref{tab:examples} illustrates the linguistic phenomenon targeted by the corpus: each utterance contains at least one code-switched medical term (boldface), ranging from single to multiple insertions within fluent Vietnamese contexts.

\begin{table}[htbp]
\centering
\small
\renewcommand{\arraystretch}{1.15}
\setlength{\arrayrulewidth}{0.5pt}
\begin{tabularx}{\linewidth}{|>{\RaggedRight\arraybackslash}X|}
\hline
\multicolumn{1}{|c|}{\textbf{Examples for CS Utterances}} \\ \hline
Tiếp theo số ba đó là cái dạng \textbf{peptide} mà nó ức chế dẫn truyền thần kinh. \\ \hline
Các loại \textbf{protit} có nguồn gốc động vật có giá trị dinh dưỡng cao, còn các \textbf{protit} thực vật có giá trị dinh dưỡng thấp. \\ \hline
Các nghiên cứu cho thấy \textbf{quercetin} có thể hoạt động bằng cách ngăn chặn hoạt động của các hóa chất gây viêm trong cơ thể như \textbf{prostan} và \textbf{leukotriene}. \\ \hline
\end{tabularx}
\caption{Representative utterances with increasing numbers of code-switched medical terms (1, 2, 3).}
\label{tab:examples}
\end{table}

\begin{figure}[t]
  \centering

    \hspace*{-0.020\linewidth}
    \includegraphics[width=\linewidth]{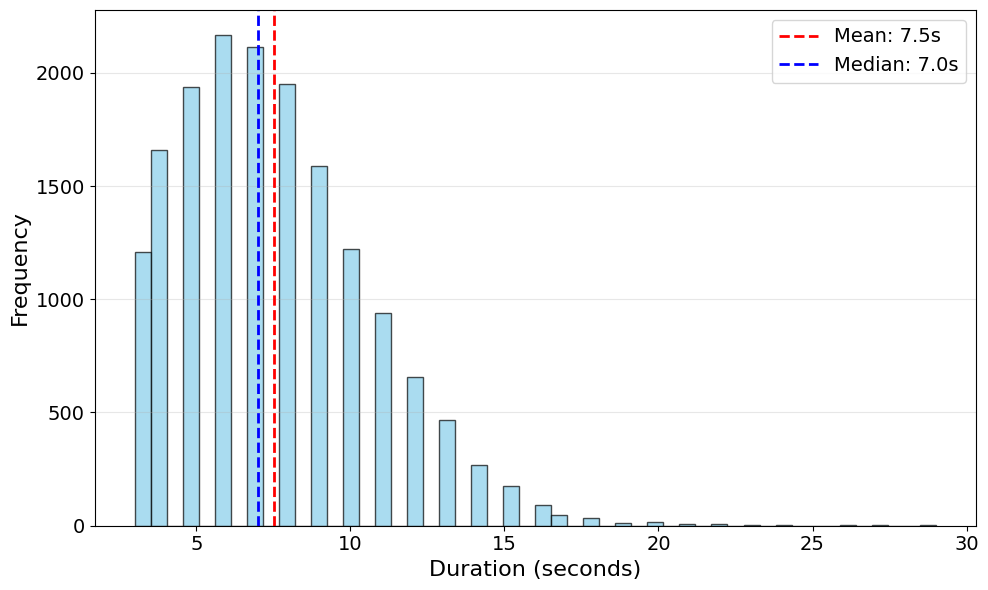}

  \caption{Histogram of utterance durations.}
  \label{fig:data_stat}
\end{figure}

The utterances are grouped into five medical topics—Medical Sciences, Pathology \& Pathogens, Treatments, Nutrition, and Diagnostics—using automatic assignment with Gemini~2.5~Pro followed by manual checks. Figure~\ref{fig:data_stat} visualizes the segment-duration distribution, Table~\ref{tab:topic_dist} reports hours and utterance counts per topic.

\begin{table}[htbp]
\centering
\small
\renewcommand{\arraystretch}{1.12}
\setlength{\arrayrulewidth}{0.5pt}
\begin{tabular}{|l|c|c|}
\hline
\textbf{Topics} & \textbf{Duration} & \textbf{\# Utterances} \\ \hline
Medical Sciences        & 16.33 h & 7{,}477 \\ \hline
Pathology \& Pathogens  & 10.27 h & 4{,}937 \\ \hline
Treatments              & 3.82 h  & 1{,}976 \\ \hline
Nutrition               & 2.14 h  & 1{,}155 \\ \hline
Diagnostics             & 2.02 h  & 1{,}031 \\ \hline
\end{tabular}
\caption{Per-topic distribution by total duration and number of utterances.}
\label{tab:topic_dist}
\end{table}

Overall, the dataset contains 16{,}576 utterances (34.6 hours). Segment lengths range from 3–29\,s and follow a unimodal, mildly right-tailed distribution (Fig.~\ref{fig:data_stat}); the mean is slightly above the median and most segments are under about 12\,s, which helps reduce padding and improves batch efficiency. The topic mix is intentionally skewed to mirror real usage (Table~\ref{tab:topic_dist}): Medical Sciences contributes roughly half of the hours, Pathology \& Pathogens forms a substantial second share, and Treatments, Nutrition, and Diagnostics make up the remainder, yielding both broad scientific coverage and focused procedural or lifestyle content.

We collected 889 distinct code-switched medical terms from the dataset. The distribution is long-tailed: 160 terms appear exactly once (18.0\%), 435 appear at most five times (48.9\%), and 207 appear at least twenty times (23.3\%). This skew mirrors domain practice—many specialized items are rare—allowing evaluation on both frequent and infrequent terminology, with rare/unseen items further isolated in the Hard split.


\section{Experiments}
\subsection{Setup}
\subsubsection{Split}
We partition the corpus into four \emph{mutually exclusive} sets: \textit{Train}, \textit{Valid}, \textit{Test}, and a dedicated \textit{Hard} split. The Hard split contains only code-switched medical terms that occur once or twice in the entire collection; all occurrences of those terms are removed from Train/Valid/Test to prevent leakage. The remaining pool is divided 8:1:1 into Train/Valid/Test while preserving speaker and topic diversity (Table~\ref{tab:splits4}).

\begin{table}[htbp]
\centering
\small
\renewcommand{\arraystretch}{1.15}
\setlength{\arrayrulewidth}{0.5pt}
\begin{tabular}{|l|c|c|c|}
\hline
\textbf{Split} & \textbf{Duration} & \textbf{\# Utterances} & \textbf{CS terms} \\ \hline
Train & 24.31 h & 11{,}833 & 610 \\ \hline
Dev & 3.56 h  & 1{,}714  & 523 \\ \hline
Test  & 3.38 h  & 1{,}615  & 509 \\ \hline
Hard  & 1.38 h  & 658      & 338 \\ \hline
\end{tabular}
\caption{Dataset splits (rows are sets; columns are statistics). The Hard set is disjoint from Train/Dev/Test.}
\label{tab:splits4}
\end{table}

\begin{table*}[t]
\centering
\small
\setlength{\tabcolsep}{4.2pt}
\renewcommand{\arraystretch}{1.12}
\setlength{\arrayrulewidth}{0.5pt}

\begin{tabular}{|l|r|r|r|r|r|r|r|r|}
\hline
\multirow{2}{*}{\textbf{Models}}
& \multicolumn{4}{c|}{\textbf{Test set}}
& \multicolumn{4}{c|}{\textbf{Hard set}} \\
\cline{2-9}
 & \textbf{WER} $\downarrow$ & \textbf{CER} $\downarrow$ & \textbf{CS-WER} $\downarrow$ & \textbf{N-WER} $\downarrow$
 & \textbf{WER} $\downarrow$ & \textbf{CER} $\downarrow$ & \textbf{CS-WER} $\downarrow$ & \textbf{N-WER} $\downarrow$ \\
\hline
MMS                    & 58.30 & 32.11 & 68.44 & 59.80 & 64.19 & 35.51 & 73.25 & 65.85 \\
\hline
wav2vec2-base-vi            & 44.74 & 25.48 & 69.05 & 46.56 & 58.21 & 30.72 & 73.08 & 61.11 \\
\hline
Whisper-Small          & 50.03 & 34.34 & 61.25 & 51.35 & 69.20 & 44.72 & 66.75 & 71.81 \\
\hline
PhoWhisper-Small       & 36.31 & 22.64 & 62.55 & 37.59 & 46.27 & 28.11 & 66.01 & 48.13 \\
\hline
Whisper-Large-v3       & 34.47 & 24.61 & \textbf{46.69} & 35.26 & 39.35 & 27.46 & \textbf{50.10} & 40.37 \\
\hline
PhoWhisper-Large       & 31.24 & \textbf{19.25} & 55.05 & 32.36 & 37.37 & \textbf{23.02} & 57.37 & 38.67 \\
\hline
VietASR                & \textbf{27.56} & 20.38 & 58.38 & \textbf{28.25} & \textbf{34.43} & 25.28 & 60.71 & \textbf{35.34} \\
\hline

\end{tabular}
\caption{Zero-shot baselines with different models.}
\label{tab:zero_onewide_rules}
\end{table*}

\begin{table*}[t]
\centering
\small
\setlength{\tabcolsep}{4.2pt}
\renewcommand{\arraystretch}{1.12}
\setlength{\arrayrulewidth}{0.5pt}
\begin{tabular}{|l|r|r|r|r|r|r|r|r|}
\hline
\multicolumn{1}{|c|}{\textbf{Methods}} 
& \multicolumn{4}{c|}{\textbf{Test Set}} 
& \multicolumn{4}{c|}{\textbf{Hard Set}} \\
\cline{2-9}
\multicolumn{1}{|c|}{}
& \textbf{WER} $\downarrow$ & \textbf{CER} $\downarrow$ & \textbf{CS-WER} $\downarrow$ & \textbf{N-WER} $\downarrow$
& \textbf{WER} $\downarrow$ & \textbf{CER} $\downarrow$ & \textbf{CS-WER} $\downarrow$ & \textbf{N-WER} $\downarrow$ \\
\hline
Frozen            & 36.31 & 22.64 & 62.55 & 37.59 & 46.27 & 28.11 & 66.01 & 48.13 \\
\hline
DV     & 36.29 & 22.63 & 62.46 & 37.57 & 46.07 & 27.95 & 66.01 & 47.94 \\
\hline
RS      & 35.60 & 23.68 & 60.07 & 36.54 & 43.89 & 29.64 & 61.52 & 45.10 \\
\hline
AdaCS             & 39.40 & 28.62 & 32.91 & 34.49 & 50.29 & 37.23 & 67.43 & 48.12 \\
\hline
LoRA              & 27.13 & 18.01 & 30.26 & 27.90 & 37.27 & 24.55 & 60.71 & 38.69 \\
\hline
LoRA + AdaCS      & 30.85 & 19.52 & 30.14 & 27.90 & 40.93 & 26.90 & 60.92 & 38.81 \\
\hline
AG                & \textbf{23.67} & \textbf{14.73} & \textbf{19.50} & \textbf{24.40} & \textbf{33.73} & \textbf{21.48} & \textbf{57.29} & \textbf{34.80} \\
\hline
AG + AdaCS        & 25.82 & 15.52 & 20.86 & 25.19 & 35.00 & 22.21 & \textbf{57.29} & 34.88 \\
\hline
\end{tabular}
\caption{Fine-tuning results on PhoWhisper-small across methods. }
\label{tab:phowhisper_finetune}
\end{table*}

\subsubsection{Metrics}
Following prior work on Vietnamese code-switching ASR, we report WER and CER together with CS-WER and N-WER to disentangle accuracy on code-switched spans from the rest of the transcript \cite{adacs}. Concretely, \emph{CS-WER} is the word error rate computed only on tokens inside code-switched regions, \emph{N-WER} is the word error rate restricted to tokens that do not require normalization (i.e., outside CS spans), and \emph{WER} is computed over the full, normalized output sequence. We compute all metrics on the mixed \textit{Test} set and report the \textit{Hard} set separately to diagnose generalization to rare/unseen medical terms.

\subsubsection{Models \& Methods}
To situate our benchmark, we report zero-shot results from representative systems along the above axes: MMS (multilingual CTC) and its Vietnamese-only counterpart wav2vec2-base-vi (self-supervised pretraining plus VLSP fine-tuning) \cite{pratap2024scaling,Thai_Binh_Nguyen_wav2vec2_vi_2021}; Whisper (Small, Large-v3; multilingual encoder–decoder) and the Vietnamese-adapted PhoWhisper (Small/Large) \cite{radford2023robust,PhoWhisper}; and VietASR (Zipformer with ASR-biased self-supervision) \cite{zhuo25}. This set balances multilingual and monolingual training and decoder paradigms while avoiding architectural redundancy; fuller background appears in Section~\ref{sec:relatedwork}.

We probe adaptation strategies on a common backbone and choose \textit{PhoWhisper-small} as the base: Whisper-style models are the prevailing substrate for recent CS-ASR, and PhoWhisper offers a Vietnamese-optimized instantiation of practical size. We group methods into four families. (i) \emph{Contextual biasing in the decoder}: \textit{Dynamic Vocabulary} (DV) extends the output inventory at inference so that each entry in a bias list is represented as a single token, enabling phrase-level biasing without external LMs \cite{sudo2024contextualized, sudo25b_interspeech}; and \textit{Rank \& Selection} (RS) ranks a large bias list with an auxiliary scorer and forwards only the top-$k$ items to the decoder for scalable contextualization \cite{hou25_interspeech}. (ii) \emph{Post-processing with contextualization}: \textit{AdaCS} adds a bias-attention normalization module to identify and normalize code-switched phrases given an external list \cite{adacs}. (iii) \emph{Parameter-efficient adapters}: \textit{LoRA} inserts low-rank adapter matrices into transformer blocks to fine-tune a small parameter subset \cite{hu2022lora}. (iv) \emph{LID-guided adaptation}: \textit{Attention Guide} (AG) selects attention heads indicative of language identity and guides them during adaptation to handle switches; prior work reports strong results on SEAME using this approach \cite{aditya2024attention, agro2025code}. Because AdaCS operates after decoding, we also evaluate hybrids (\textit{LoRA+AdaCS}, \textit{AG+AdaCS}). For all contextual-biasing methods (DV, RS, AdaCS), the bias list is built from the code-switched medical terms present in the corresponding split and used when decoding that split (train/test/hard), ensuring split-consistent contextualization; for AG, we employ bilingual Vietnamese–English prompts to reflect the intended CS setting.

\begin{table*}[t]
\centering
\small
\setlength{\tabcolsep}{4.2pt}
\renewcommand{\arraystretch}{1.12}
\setlength{\arrayrulewidth}{0.5pt}
\begin{tabular}{|l|r|r|r|r|r|r|r|r|}
\hline
\multicolumn{1}{|c|}{\textbf{Methods}}
& \multicolumn{4}{c|}{\textbf{Test Set}}
& \multicolumn{4}{c|}{\textbf{Hard Set}} \\
\cline{2-9}
\multicolumn{1}{|c|}{}
& \textbf{WER} $\downarrow$ & \textbf{CER} $\downarrow$ & \textbf{CS-WER} $\downarrow$ & \textbf{N-WER} $\downarrow$
& \textbf{WER} $\downarrow$ & \textbf{CER} $\downarrow$ & \textbf{CS-WER} $\downarrow$ & \textbf{N-WER} $\downarrow$ \\
\hline
Whisper-Small    & 50.03 & 34.34 & 61.25 & 51.35 & 69.20 & 44.72 & 66.75 & 71.81 \\
\hline
PhoWhisper-Small & 36.31 & 22.64 & 62.55 & 37.59 & 46.27 & 28.11 & 66.01 & 48.13 \\
\hline
Whisper-Small (LoRA)      & 39.96 & 26.41 & 33.24 & 41.35 & 46.31 & 31.91 & 60.92 & 47.78 \\
\hline
PhoWhisper-Small (LoRA)   & 27.13 & 18.01 & 30.26 & 27.90 & 37.27 & 24.55 & 60.71 & 38.69 \\
\hline
Whisper-Small (AG)   & 24.54	& 15.05 & 20.86 & 25.23 & 35.26 & 22.07	& 56.84 & 36.36 \\
\hline
PhoWhisper-Small (AG)  & \textbf{23.67} & \textbf{14.73} & \textbf{19.50} & \textbf{24.40} &\textbf{33.73} & \textbf{21.48} & \textbf{57.29} & \textbf{34.80} \\
\hline
\end{tabular}
\caption{Effect of fine-tuning approaches by LoRA \& Attention Guide on multilingual (Whisper-Small) models vs. monolingual (PhoWhisper-Small).}
\label{tab:mono}
\end{table*}

\subsection{Zero-shot results}

As shown in Table~\ref{tab:zero_onewide_rules}, there is a stable split between sentence-level accuracy and code-switched spans across both test and hard sets. Within each model pair, Vietnamese-optimized systems reduce utterance errors (WER/CER/N\mbox{-}WER) relative to their multilingual counterparts: wav2vec2-base-vi improves over MMS, PhoWhisper-Small over Whisper-Small, and PhoWhisper-Large over Whisper-Large-v3. Overall, VietASR is strongest on sentence-level metrics (best WER and N\mbox{-}WER on both splits), and PhoWhisper-Large attains the lowest CER among the large-capacity models. These outcomes are consistent with extensive Vietnamese-only pretraining and targeted fine-tuning that better capture matrix-language phonotactics and style.

On the code-switched regions, the pattern reverses at the high-capacity end: Whisper-Large-v3 delivers the lowest CS-WER on both splits, outperforming VietASR and the PhoWhisper variants (e.g., on the test set it leads by a clear margin). At smaller scale the gap narrows and can be comparable, but PhoWhisper-Small still retains its advantage on overall WER/CER/N\mbox{-}WER. Taken together, the results reinforce a common trade-off also observed on external CS benchmarks: monolingual or Vietnamese-adapted models dominate sentence-level accuracy, whereas broad multilingual exposure improves recognition of embedded English “islands” within Vietnamese utterances \cite{decm,agro2025code}.

\subsection{Fine-tuning results}

Results of finetuning across methods on PhoWhisper-small (Table~\ref{tab:phowhisper_finetune}) show a consistent pattern. Decoder-side contextual methods (DV, RS) provide only modest changes relative to the frozen model, while AdaCS sharply reduces CS-WER but can raise overall WER \& CER in isolation—typical of precision–recall trade-offs when bias spans are sparse or noisy. In contrast, adapter-based fine-tuning improves all metrics, with AG yielding the strongest overall and hard-set scores and LoRA a solid second. Combining adapters with post-processing further stabilizes performance (LoRA+AdaCS, AG+AdaCS), preserving low CS-WER while recovering sentence-level accuracy. Taken together, Vietnamese medical CS benefits more from parameter-efficient adaptation—especially LID-guided AG—than from decoder-only contextualization, while contextual normalization remains a useful complement for difficult terms.

To compare monolingual and multilingual initializations under the same adaptations, Table~\ref{tab:mono} show experiments across two strongest adaptation strategies-LoRA and Attention Guide (AG) on Whisper-small and PhoWhisper-small. Both gain markedly, but PhoWhisper ends up stronger overall and on most CS metrics. With LoRA, PhoWhisper cuts CS-WER by roughly half—more than the reduction observed for Whisper—and also achieves lower WER/CER/N-WER on the test split while keeping an edge on the hard split. AG pushes performance further: PhoWhisper attains the lowest test errors, with CS-WER slightly below Whisper’s and sentence-level metrics clearly in its favor; on the hard split, CS-WER is comparable across models, but PhoWhisper retains better WER and N-WER. In short, after fine-tuning, the Vietnamese-optimized backbone learns code-switched medical terms more effectively and delivers consistently stronger accuracy than its multilingual counterpart.

\begin{table}[htbp] 
\centering
\small
\setlength{\tabcolsep}{6pt}
\renewcommand{\arraystretch}{1.12}
\setlength{\arrayrulewidth}{0.5pt}
\begin{tabular}{|l|r|r|r|}
\hline
\multicolumn{1}{|c|}{\textbf{Topics}} &
\multicolumn{1}{c|}{\textbf{Frozen}} &
\multicolumn{1}{c|}{\textbf{LoRA}} &
\multicolumn{1}{c|}{\textbf{AG}} \\
\hline
Medical Sciences        & 68.17 & 41.28 & \textbf{34.28} \\
\hline
Pathology \& Pathogens  & 63.86 & 37.07 & \textbf{26.92} \\
\hline
Treatments              & 73.74 & 34.30 & \textbf{27.44} \\
\hline
Nutrition               & 53.82 & 16.92 & \textbf{13.85} \\
\hline
Diagnostics             & 54.07 & 52.01 & \textbf{44.40} \\
\hline
\end{tabular}
\caption{CS-WER by different methods based on PhoWhisper-Small divided by topics.}
\label{tab:topic_cs_wer}
\end{table}

Moreover, Table~\ref{tab:topic_cs_wer} shows consistent gains across topics after adaptation. Treatments is the most error-prone category in the frozen model but no longer the worst once fine-tuned, while Nutrition starts easiest and remains so. Medical Sciences, the largest and most diverse topic, continues to be comparatively challenging even after adaptation, and Diagnostics also trails the middle group. Across all topics, AG yields the lowest CS-WER and LoRA is a close second, indicating that adapter-based methods substantially narrow cross-topic gaps and shift the error peak away from Treatments, though broad scientific content still stresses the model.

\section{Conclusion}
In this paper, we introduced \textbf{ViMedCSS}, the first publicly available benchmark dataset for Vietnamese medical code-switching (CS) speech, containing 34.6 hours and 16,576 utterances. This resource addresses a critical gap in ASR development, as we demonstrated that standard models struggle to recognize English medical terms embedded in Vietnamese. Our zero-shot experiments revealed a clear performance trade-off: multilingual models like Whisper-Large-v3 excel at recognizing English CS terms, whereas Vietnamese-optimized models like VietASR are superior for the surrounding Vietnamese text, resulting in lower overall word error rates.

To resolve this, we investigated several fine-tuning strategies, finding that parameter-efficient adaptation offers the most effective solution. Notably, applying the Attention Guide (AG) adaptation method to a Vietnamese-specialized model (PhoWhisper-Small) yielded the best performance, significantly reducing errors on both CS terms and general speech. This work not only provides a valuable dataset for the community but also identifies a clear and effective fine-tuning approach for building robust, domain-specific ASR models in low-resource and code-switching contexts. Future work can leverage this benchmark to explore further enhancements in contextual biasing and model architecture.

\section{Acknowledgments}
This research was supported by the VinUniversity Cross-College Research Grant (Grant ID: VUNI.2324.CC06).
\section{Ethics Statement}
The data were collected from a publicly available source, YouTube. The content extracted from this source is used for research purposes only, and it does not contain any private information about patients.

\nocite{*}
\section{Bibliographical References}\label{sec:reference}

\bibliographystyle{lrec2026-natbib}
{
\renewcommand{\bibsection}{}
\bibliography{lrec2026-example}
}
\label{lr:ref}
\bibliographystylelanguageresource{lrec2026-natbib}
\bibliographylanguageresource{languageresource}

\end{document}